\documentclass[letterpaper, 10 pt, conference]{ieeeconf}  

\IEEEoverridecommandlockouts                              

\usepackage{graphics} 
\usepackage{subcaption}
\usepackage{graphicx}
\usepackage{mwe}
\usepackage{epsfig} 
\usepackage{amsmath} 
\usepackage{amsfonts} 

\title{\LARGE \bf
From Scratch to Sketch: Deep Decoupled Hierarchical Reinforcement Learning for Robotic Sketching Agent
}

\author{Ganghun Lee$^{1}$, Minji Kim$^{2}$, Minsu Lee$^{4}$ and Byoung-Tak Zhang$^{3, 4}$
\thanks{*This work was partly supported by the NRF of Korea (2021R1A2C10-10970/20\%), %
the IITP (2018-0-00622-RMI/20\%, %
2019-0-01371-BabyMind/20\%, %
2015-0-00310-SW.StarLab/10\%, %
2021-0-02068-AIHub/10\%), %
and the KIAT (P0006720-ILIAS/10\%) grants funded by the Korean government, %
and the CARAI (UD190031RD/10\%) grant funded by the DAPA and ADD.}
\thanks{$^{1}$Interdisciplinary Program in Cognitive Science, Seoul National University, Seoul, Korea}%
\thanks{$^{2}$Interdisciplinary Program in Neuroscience, Seoul National University}%
\thanks{$^{3}$Dept. of Computer Science and Engineering, Seoul National University}%
\thanks{$^{4}$AIIS, Seoul National University}%
\thanks{{\tt\small \{khlee, mjkim, mslee, btzhang\}@bi.snu.ac.kr}}
}

\begin{document}

\maketitle
\thispagestyle{empty}
\pagestyle{empty}

\begin{abstract}
We present an automated learning framework for a robotic sketching agent that is capable of learning stroke-based rendering and motor control simultaneously. 
We formulate the robotic sketching problem as a deep decoupled hierarchical reinforcement learning; two policies for stroke-based rendering and motor control are learned independently to achieve sub-tasks for drawing, and form a hierarchy when cooperating for real-world drawing.
Without hand-crafted features, drawing sequences or trajectories, and inverse kinematics, the proposed method trains the robotic sketching agent from scratch.
We performed experiments with a 6-DoF robot arm with 2F gripper to sketch doodles.
Our experimental results show that the two policies successfully learned the sub-tasks and collaborated to sketch the target images.
Also, the robustness and flexibility were examined by varying drawing tools and surfaces.
\end{abstract}

\section{Introduction}
Sketches have been successfully used to graphically demonstrate scenes, ideas, or principles as they are a natural but powerful means of expressing internal thoughts.
If we can develop robots that can sketch any idea with any drawing medium, the robotic drawing technology will induce big synergies and opportunities in everyday life by incorporating the state-of-the-art AI technologies such as natural language understanding and generative models. Through human-robot interactions, the robots can be used to express or expand humans’ ideas \cite{hri1}.
Moreover, the robots can be utilized to teach children how to draw sketches, express their imaginations, or visualize their internal concepts.

\begin{figure}[t]
  \centering
  \includegraphics[scale=0.5]{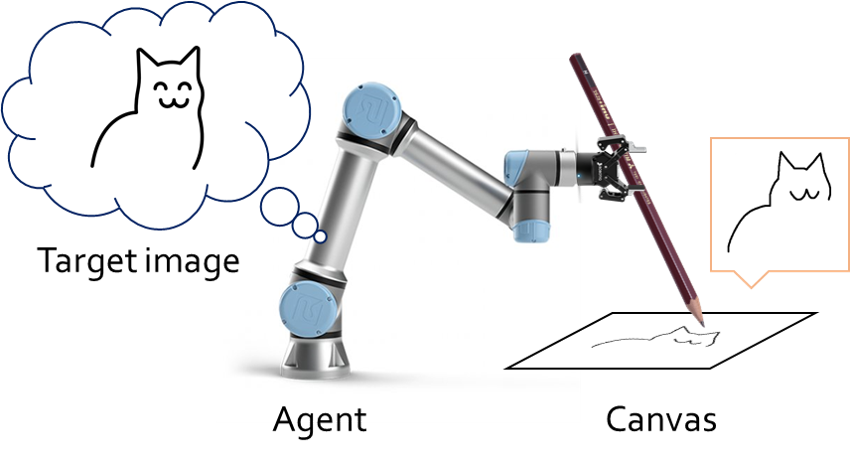}
  \caption{Task of our agent. A target image is given, then the agent generates a sketch of the target on the real canvas.}
  \label{task}
\end{figure}

Because humans have been drawing through practicing and playing since childhood, adults usually can make simple drawings without effort. 
However, for robots, drawing is a challenging task due to the complicated skills it requires.
In this paper, we address the problem of developing an automated learning framework for robotic sketching agent (Fig. \ref{task}). 
We suggest sketching requires mainly two skilled behaviors.
First is an in-mind intelligence to decide in what order to put strokes on the canvas.
Second is an embodiment control to accomplish each stroke.
The high-level intelligence decides and passes the stroke commands to the low-level intelligence which is dedicated to carrying out the commands.
Overall, these two behaviors cooperate in a hierarchical manner to materialize an imagination into visual representation.

The task for the high-level intelligence can be described as stroke-based rendering (SBR).
SBR is a method of image recreation that places discrete drawing elements such as paint strokes or stipples on canvas \cite{sbr}.
However, many recent studies on SBR \cite{sketchrnnquickdraw, strokenet, learningtopaint, spiral, cml} are not sufficient to directly deploy in the real-world.
Because most SBR models are developed on the virtual rendering assumptions, their renderings are hard to implement with the real-world drawing tools.

Even if the SBR stroke setup is well-established, control problem for the low-level intelligence still remains as another challenge.
Although some of SBR studies considered robots as components for demonstration, their motor controls were planned by using inverse kinematics or manual controls \cite{nao}.
The movements hence were strict and lacking spontaneity, and usually maintained the drawing tools vertically \cite{automaticpaintingreview}.
But thanks to recent advances in robotics, especially with machine learning, there is room for teaching robots to draw in human-like manner.
Several studies indeed already took machine learning techniques for robots to perform various kinds of artistic strokes~\cite{artisticstyle, automaticpaintingreview}.
However, training for high-level stroke planning, SBR, is absent from them.

\begin{figure*}[t]
  \centering
  \includegraphics[scale=0.51]{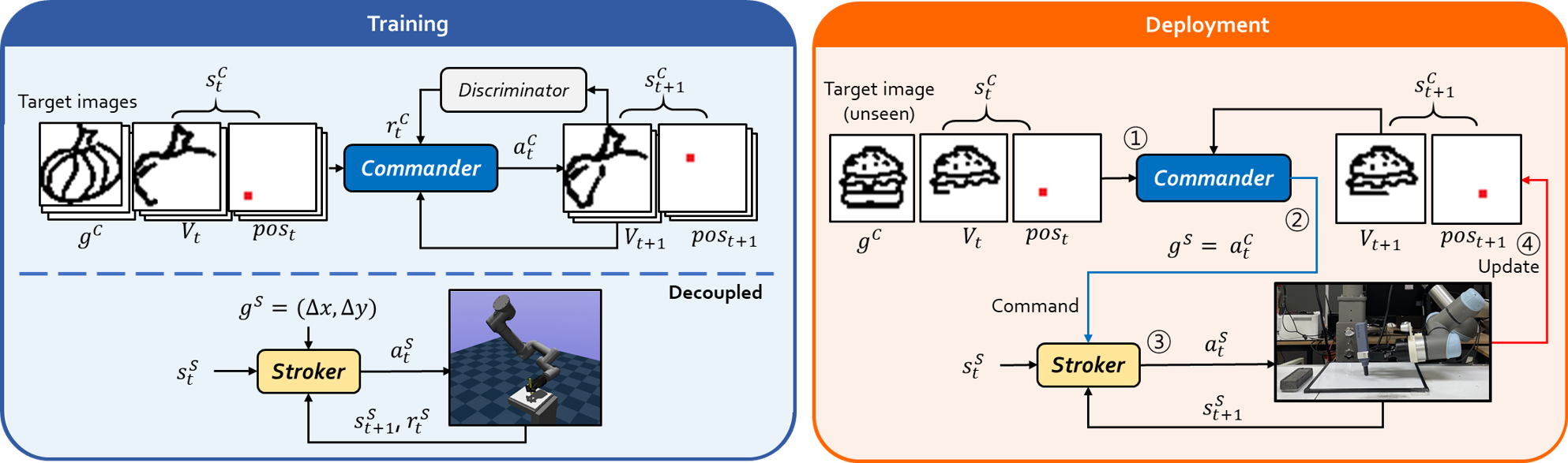}
  \caption{Overview of our deep decoupled hierarchical RL for sketch agent. In training, Commander and Stroker learn their task independently with RL (left). In deployment, the two collaborate and work hierarchically (right). First, given a target image which is unseen in training, then Commander decides its action based on the goal, the current imaginary canvas, and the pen position. Second, the action of Commander is transmitted to Stroker. Third, Stroker decides the control values for robot arm, based on the current joint angles. Finally, robot arm executes Stroker's action to draw on the real canvas and sends pentip position to update Commander's imaginary canvas. Based on the updated imaginary canvas, Commander decides the action again, and this procedure is repeated until the sketching completes.}
  \label{architecture}
\end{figure*}

In this paper, we present an automated learning framework that integrates high-level SBR task and low-level control task for robot drawing through reinforcement learning (RL).
We formulate the problem as deep decoupled hierarchical reinforcement learning; the high- and low-level policies are independently trained to achieve sub-tasks for drawing, and form a hierarchy when collaborating for real-world drawing.
We would like to mention that using RL in motor control for drawing is also a novel point of our work.
Even though RL is one of the major machine learning technologies and has produced promising results in robotics domain \cite{rlrobot1, rlrobot2, rlrobot3, rlrobot4}, the usage of RL for low-level control of real-world robots for drawing has not yet been conducted.
By using RL in both stages of the learning process, we have the advantage of not requiring any stroke sequences or trajectory samples for training our sketching agent.
That is, any kinds of SBR methods and any forms of robot joints can be directly trained without specific priors.
Our key contributions are summarized by the following:
\begin{itemize}
\item Effectiveness: The proposed formulation integrates two sub-tasks for robotic sketch automation.
\item Scalability: RL-based training scheme without expensive data enables general robots to learn to sketch.
\item Robustness and stability: Imaginary canvas synchronizing with real-world canvas allows robots to sketch without camera, regardless of drawing conditions.
\end{itemize}

\section{Related Work}
\subsection{Stroke-Based Rendering}
The goal of SBR is to find methods to reproduce a target image by placing a number of discrete visual elements such as paintbrush strokes, pen-and-ink, or stipples on canvas \cite{sbr}.
Traditional SBR methods used optimization algorithms or greedy search on every single step, but these algorithms are too slow and require heavy human intervention.

Recent works used advanced machine learning techniques to improve the stroke decomposition of images.
SketchNet \cite{sketchrnnquickdraw} and StrokeNet \cite{strokenet} used recurrent neural networks (RNN) \cite{lstm} with supervised learning to construct stroke-based drawings.
However, those methods require significant amount of stroke sequence data for each drawing.
Instead, Learning to paint \cite{learningtopaint}, SPIRAL \cite{spiral}, and SPIRAL++ \cite{spiral++} adopted RL techniques with adversarial approach to train drawing agent without any stroke sequences.
Above studies have made great progress in SBR research by incorporating deep learning techniques.
Nevertheless, to the best of our knowledge, there are still no studies which have considered RL mechanism for both SBR and low-level robot control simultaneously.

\subsection{Robotic Drawing}
The earliest fully automated drawing machine is known as automata in the 1970s that is a small doll that can sketch pre-encoded images using a clockwork-like movement mechanism \cite{automata}.
Recently, general robots or manipulator robots with high DoFs have been used for drawing tasks.
Paul the robot \cite{paultherobot} is a robotic hand-eye system that produces observational portrait drawing by mimicking artists’ drawing skills and techniques.
~\cite{nao} demonstrated the sketch drawing capability of NAO humanoid robot.
Above studies take inverse kinematics for robot control and use hand-crafted visual calibration techniques.

Some other researches took learning approach for robot drawing.
\cite{teachrobottodrawbextor} presented a manipulator-based robotic drawing study that can infer a sequence of commands for drawing a character given a bit-mapped image of the character.
\cite{artisticstyle} proposed a method to learn brushstrokes from human artists.
While the above studies showed meaningful progress in robot drawing, they require demonstration data from human which is costly to construct.

Unlike these approaches, our framework does not require any hand-crafted rules, stroke-order demonstrations, or cameras for checking drawing status.
Our proposed approach can train a robot to sketch based on stroke decision and motor control from scratch using a decoupled version of hierarchical reinforcement learning algorithm.

\subsection{Hierarchical Reinforcement Learning}
Hierarchical reinforcement learning (HRL) decomposes a long-horizon reinforcement learning problem into a hierarchy of sub-tasks such that high-level parent tasks invoke low-level child tasks as if they are primitive actions.
Such sub-tasks may both be easier to learn and lead to more structured and improved exploration \cite{hrl}.
Several HRL studies have demonstrated success in solving some difficult RL tasks \cite{hrlcs, hrl, hiro, hac}.
While our approach follows the policy structure of HRL, it should be noted that we decoupled the training of hierarchical policies to achieve each level of tasks.

\section{Method}
Sketch task requires hierarchical skills that comprise stroke decision and motor control.
We define two policies for both skills: \textit{Commander} and \textit{Stroker} as illustrated in Fig. \ref{architecture}.
Commander is trained to reconstruct the given goal image by determining sequential strokes on imaginary white canvas.
Then, the stroke command generated by Commander becomes a goal for Stroker that is trained to implement the stroke in the physical world.
We train Commander in imaginary canvas environment without physical devices, whereas Stroker is trained with robot arm in simulated environment. 
Both policies are trained with RL, without requiring any data of specific stroke sequences or arm trajectories.

In this section we first introduce notations and summarize the standard RL framework.
Next, we propose a formulation of deep decoupled hierarchical reinforcement learning for robot drawing task.
Finally, detailed training schemes of each policy are described.

\subsection{Background}
We address policy learning in continuous action spaces used for both policies.
We consider a finite-horizon Markov decision process (MDP) defined by the tuple ($\mathcal{S}$, $\mathcal{A}$, $p$, $r$, $\gamma$), where the state space $\mathcal{S}$ and the action space $\mathcal{A}$ are continuous, and the state transition probability $p(s_{t+1}|s_t,a_t)$ represents the probability density of the next state $s_{t+1} \in \mathcal{S}$ given the current state $s_t \in \mathcal{S}$ and action $a_t \in \mathcal{A}$.
The environment emits a reward $r$ on each transition.
The objective in standard RL is to maximize the expected sum of discounted rewards $\sum_t\mathbb{E}_{(s_t,a_t)\sim\rho_\pi}[\gamma^t r(s_t,a_t)]$ designed to achieve a single task via training the policy $\pi(a_t|s_t)$, where $\rho_\pi(s_t)$ and $\rho_\pi(s_t,a_t)$ are the state and state-action marginals of the trajectory distribution induced by a policy $\pi(a_t|s_t)$, and the discount factor $\gamma\in [0,1]$ describes how valuable the future rewards are.

In our case, we consider goal-conditioned RL which is commonly modified version of RL for multiple tasks \cite{goalrlold, goalrl1, goalrl2, goalrl3, goalrlplanning}; the policy $\pi(a_t|s_t, g)$ is conditioned on a goal $g \in \mathcal{G}$ trained to take proper action to achieve the given goal, where the goal space $\mathcal{G}$ is continuous.
The reward $r(s_t,|a_t,g)$ is also conditioned by the goal.

\subsection{Deep Decoupled Hierarchical Reinforcement Learning}
We define deep decoupled hierarchical reinforcement learning (DHRL) for our task.
Brief overview about our architecture is illustrated in Fig. \ref{architecture}.
The ultimate goal $g^C$ sampled from Commander's goal space $\mathcal{G}_C$ conditions Commander's policy $\pi_C(a^C_t|s^C_t,g^C)$, where $a^C_t \in \mathcal{A}_C$ and $s^C_t \in \mathcal{S}_C$ are action and state from Commander's action space $\mathcal{A}^C$ and state space $\mathcal{S}^C$.
The action taken by Commander becomes Stroker's goal $a^C_t=g^S \in \mathcal{G}_S$, which then conditions Stroker's policy $\pi_S(a^S_t|s^S_t,g^S)$, where $a^S_t \in \mathcal{A}_S$ and $s^S_t \in \mathcal{S}_S$ are action and state from Stroker's action space $\mathcal{A}^S$ and state space $\mathcal{S}^S$.
The action of Commander is the same as the goal for Stroker, and hence Stroker's goal space is the same as Commander's action space $\mathcal{A}_C=\mathcal{G}_S$.

The chain of Commander's action and Stroker's goal might be merged and trained at once as in HRL, but we find that applying such approach to our task is unrealistic because of the substantial gap between imaginary stroke planning and physical joint controls.
Moreover, if trained together, they form mutual dependency which makes replacing the trained policy with another unavailable, and makes it hard to track the training process.
Rather, we consider decoupled training for both policies to ensure modularity and manageability.

\begin{figure}[t]
  \centering
  \includegraphics[scale=0.45]{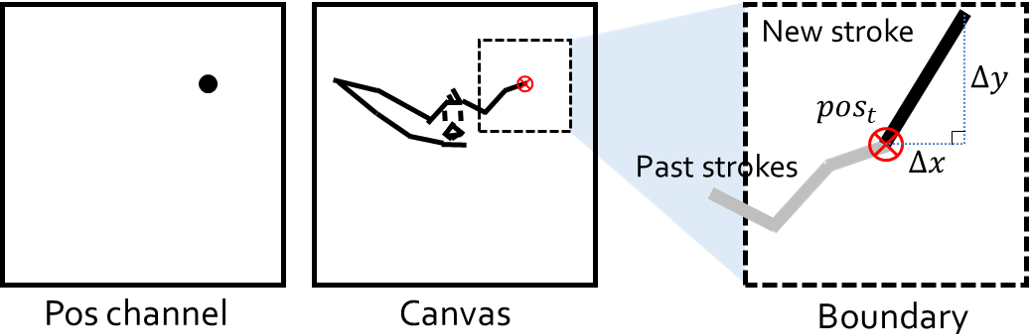}
  \caption{Commander receives the concatenation of canvas and pos channel to decide an action. Pos channel indicates where the pen is located. Commander's action determines the end point of the new straight line which can be placed within the fixed length square boundary regarding the current pen position as starting point.}
  \label{commander}
\end{figure}

\subsection{Commander}
For Commander, a complete target image $I$ is given as a goal $g^C$.
Starting from the imaginary white canvas $V_0$, Commander decides the first action $a^C_0$ which determines where to put the first stroke.
Once the stroke is drawn, Commander observes the next canvas $V_1$ with the stroke drawn and takes successive steps $\{a^C_1,...,a^C_T\}$ until it completely mimics the target image.

We adopt a short straight line for one stroke which stretches from the current stroking position to the end position decided by Commander within fixed length square boundary (Fig. \ref{commander}).
Commander's action is defined as $a^C_t = (down, \Delta x, \Delta y)$, where $down \in [0, 1]$ is a pen-down indicator.
If $down > 0.5$, stroke is drawn and if $down \leq 0.5$, pen is lifted up only for changing position.
$\Delta x, \Delta y \in [0,1]$ denotes the change in each axis position, where values 0 and 1 denote leftmost/bottom and rightmost/top position within the square boundary, respectively.
Commander's policy $\pi_C(a^C_t|s^C_t,g^C)$ takes an action based on its observation at every timestep $t$.
It observes the current state $s^C_t = (V_t, pos_t)$ and the goal $g^C = I$, where $V_t$ is current canvas, $pos_t=(x,y)$ is current pen position, and $I$ is the complete target image.

To successfully train Commander, its strokes need to be evaluated.
Previous SBR works with RL took an adversarial approach inspired by GAN \cite{gan} that trains a discriminator for similarity measurement between the target image and Commander's drawing.
We adopt the same setting for Commander's reward function as follows:
$$
r^C_t = d(V_t, I) - d(V_{t-1}, I). \eqno{(1)}
$$
Here $d(V_t, I)$ stands for the similarity between Commander's current drawing $V_t$ and target image $I$ calculated by the discriminator $d$.
If the reward is positive, it means that Commander's stroking action $a^C_{t-1}$ that made transition from $V_{t-1}$ to $V_t$ helped to increase the similarity.
Following \cite{spiral++}, we adopt SNGAN \cite{sngan} for discriminator to stabilize training.
Discriminator $d$ is simultaneously trained with Commander.

With regard to the stroke style, definition of action for Commander should be carefully designed.
Previous works usually represent a stroke in absolute Cartesian coordinates system.
In particular, \cite{learningtopaint, spiral++, artisticstyle} adopt Bézier curve, and use starting coordinate, medium coordinate, end coordinate, and other specific parameters such as colors and thickness to represent a stroke.
Fine-grained stroke like above is useful to express unique styles on the imaginary canvas, where the strokes are rendered virtually.
However, complicated stroke is not a best option when a stroke is physically rendered with real drawing tools like pen or board markers, as the stroke determined by the large parameter space is such a burden for the low-level physical controller to search.
For this reason, we limited Commander's single stroke to be simple enough for Stroker.
This straightforward structure might degrade aesthetic aspect of single stroke, but allows more freedom in learning creative drawing skills because modeled strokes like Bézier curve are eventually equal to the sum of our basic strokes.

\subsection{Stroker}
Stroker, equipped with a physical robot arm, is a command executor holding a pen to directly interact with the canvas in the real world.
As the commands generated by trained Commander are relative to the current pen position, Stroker needs to learn how to move the pen from its position.
Once the real stroke is drawn, Stroker conveys the changed pen position to the Commander to update the imaginary canvas.
In this way, Commander can track Stroker's drawing without camera.
Then Commander observes the updated imaginary canvas and generates the next command to pass to Stroker again.
Repeating this process, our agent can sketch the given target image.

\begin{figure}[t]
    \centering
    \begin{subfigure}[b]{0.23\textwidth}
        \centering
        \includegraphics[width=\textwidth]{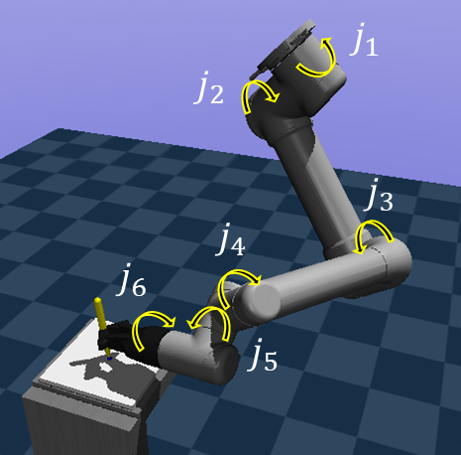}
        \caption[]%
        {{\small Joints of robot arm}}    
    \end{subfigure}
    \hfill
    \begin{subfigure}[b]{0.235\textwidth}  
        \centering 
        \includegraphics[width=\textwidth]{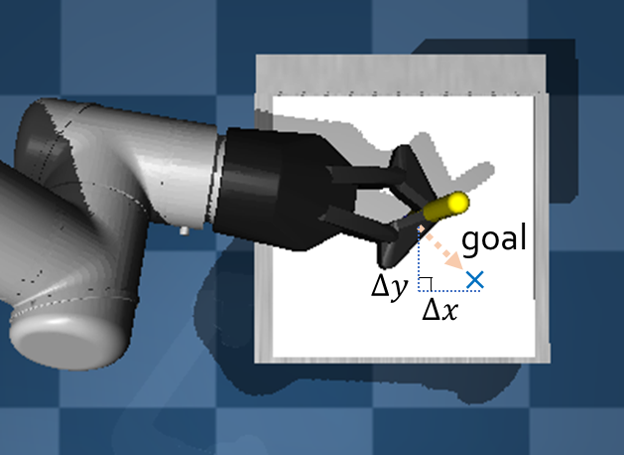}
        \caption[]%
        {{\small Goal for Stroker}}    
    \end{subfigure}
    \caption[]
    {\small (a) Illustration of robot arm with 6-DoF joints. (b) Goal for Stroker is same as the action of Commander. Stroker controls its joint angles to get pentip reached the goal position.} 
    \label{fig:stroker}
\end{figure}

Stroker's policy $\pi_S(a^S_t|s^S_t,g^S)$ decides the action $a^S_t$ after observing the current state $s^S_t$ and the given goal $g^S$.
The state for Stroker is the current joint angles of the robot arm.
Given that the robot arm has six joints, the state for Stroker $s^S_t = (j_1, j_2, j_3, j_4, j_5, j_6)$ includes the current six joint angles.
The action for Stroker $a^S_t = (\Delta j_1,\Delta j_2,\Delta j_3,\Delta j_4,\Delta j_5,\Delta j_6)$ indicates the changes in each joint angle.
Stroker is trained to achieve the goal as Commander's action $a^C_t = g^S_t = (down, \Delta x, \Delta y)$.
The illustration of Stroker's state and action for the given goal is in Fig. \ref{fig:stroker}.
The concise definition for state $s^C_t$ involving only joint values ensures that any other robot arms can be simply adopted to be a Stroker.

In training session, Stroker learns independently of Commander.
Since the stroke commands are not known beforehand, Stroker should be trained to execute any command.
Therefore, we randomly sample the goals for Stroker in training.
Stroker should also execute the commands regardless of the states of Stroker itself.
For this, random initialization of joint angles would be inappropriate because the joint space for Stroker is too vast to cover, and the majority of the space yields undesirable poses for sketching.
Instead, we define 9 initial poses for Stroker to start each set of training and let it explore the joint space near the initial poses as Stroker changes its joints trying to attain the random goals.
Since immature Stroker is likely to commit undesirable behavior inducing inapposite poses or positions, Stroker's pose is randomly initialized to predefined joint values after sampling every 100 goals.

Reward $r^S_t$ is given as an evaluation of the result of Stroker's actions.
We define reward function for Stroker as below:
$$
r^S_t = -10\sqrt{(g_x-p_{t_x})^2 + (g_y-p_{t_y})^2 + 5(g_z - p_{t_z})^2}
$$
$$
+ \ 5\cfrac{[g_x\ g_y]\cdot[p_{t_x}\ p_{t_y}]}{\lVert[g_x\ g_y]\rVert\lVert[p_{t_x}\ p_{t_y}]\rVert} \ - \ \sqrt{(\gamma_g-\gamma_t)^2+(\psi_g-\psi_t)^2}, \eqno{(2)}
$$
where the first term is modified Euclidean distance between goal position $g=(g_x,g_y,g_z)$ and reached position $p_t=(p_{t_x}, p_{t_y}, p_{t_z})$, the second term is cosine similarity between $g$ and $p_t$ about $xy$ axes, and the final term is Euler angle distance between preferable global pen rotation $(\gamma_g, \psi_g)$ and reached global pen rotation $(\gamma_t, \psi_t)$.
The first term is scaled up 10 times for scale adjustment with others.
Also, the $z$ distance in the first term is scaled up 5 times since $z$ axis determines pen lifting crucial for sketching.
Even though scale of the second term is basically biggest, we scale up it 5 times since we empirically found that the direction of the pen movement is more important than the position.
The preferable pen rotation is predefined to be slightly tilted from vertical like how humans hold the pen, but to allow a slight amount of flexibility, the final term is remained low relative to others.

\begin{figure}[t]
    \centering
    \begin{subfigure}[b]{0.235\textwidth}
        \centering
        \includegraphics[width=\textwidth]{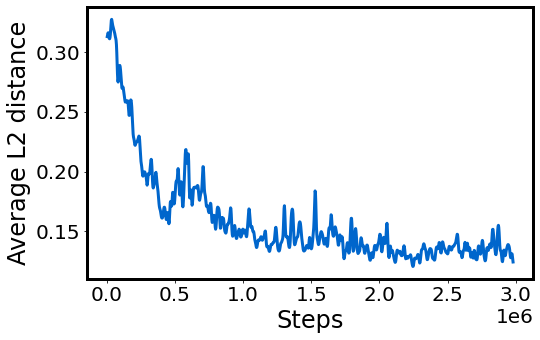}
        \caption[]%
        {{\small L2 distance}}    
    \end{subfigure}
    \hfill
    \begin{subfigure}[b]{0.235\textwidth}  
        \centering 
        \includegraphics[width=\textwidth]{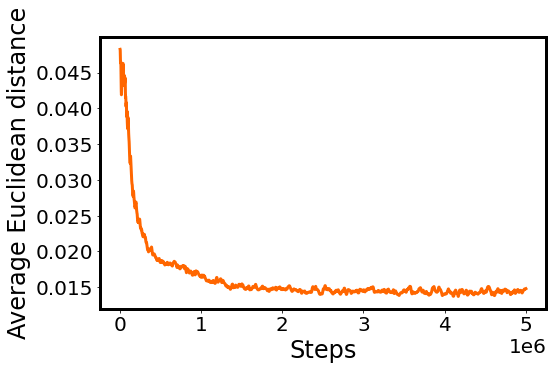}
        \caption[]%
        {{\small Euclidean distance for goal}}    
    \end{subfigure}
    \vskip\baselineskip
    \begin{subfigure}[b]{0.235\textwidth}   
        \centering 
        \includegraphics[width=\textwidth]{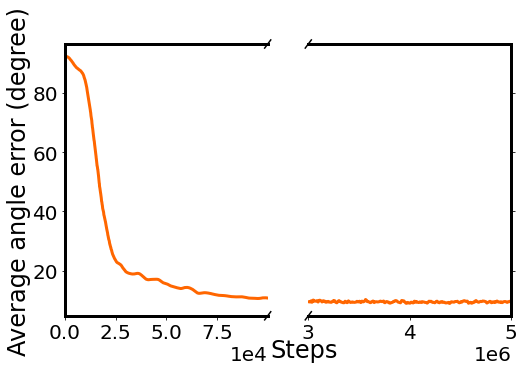}
        \caption[]%
        {{\small Angle error for goal}}    
    \end{subfigure}
    \hfill
    \begin{subfigure}[b]{0.235\textwidth}   
        \centering 
        \includegraphics[width=\textwidth]{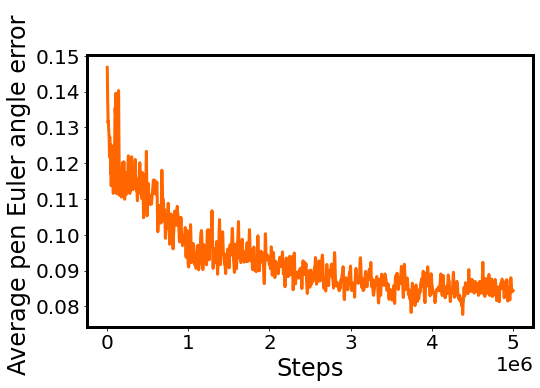}
        \caption[]%
        {{\small Euler angle error for pen}}
    \end{subfigure}
    \caption[]
    {\small Performance evaluation for Commander and Stroker during training. (a) Average L2 distance between target image and Commander's drawing. (b), (c) are average Euclidean distance and average angle error between given goal position and reached position, respectively. (d) Euler angle error between Stroker's pen rotation and preferable angle.} 
    \label{fig:learning}
\end{figure}

\begin{figure}[t]
  \centering
  \includegraphics[scale=0.7]{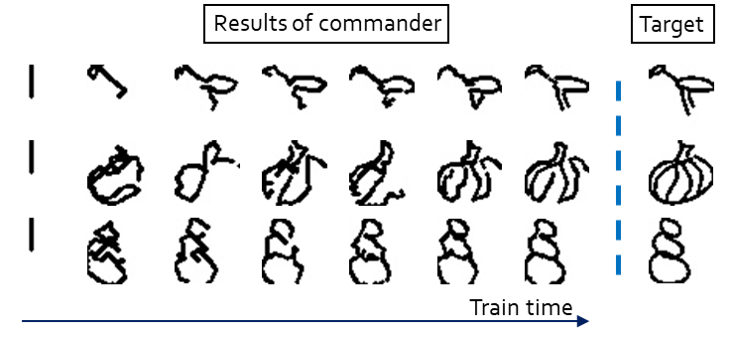}
  \caption{Performance of Commander during training. Resulting sketches in imaginary canvas gradually getting closer to the target images over training time.}
  \label{commander_train}
\end{figure}

\subsection{Deployment}
The two policies are merged to cooperate in the deployment step when the training is over (Fig. \ref{architecture} (right)).
The high-level action issued by Commander becomes the goal for Stroker, then the low-level action is issued by Stroker to directly control the robot arm.

When deploying in real scenarios, it must be considered that Stroker cannot perfectly implement the Commander's goal due to the inevitable errors to occur in Stroker and the robot.
Because Commander was trained on the assumption its high-level action is flawlessly executed, the imaginary canvas of Commander gets distorted if Commander cannot access any feedback from Stroker.
In order to compensate for the errors, a special pentip position-based synchronization technique is devised: tracking the pentip position and sending it to Commander so that Commander can update the imaginary canvas correctly.
Stroker can easily track the pentip position using forward kinematics with the issued low-level action.
Then Commander can determine where Stroker actually moved the pentip to and synchronize the imaginary canvas with the actual canvas.

\section{Experiments and Results}

\subsection{Training of Commander and Stroker}
We used \textit{Quick, Draw!} \cite{sketchrnnquickdraw} dataset, human-drawn simple doodles as the target images for Commander's training. 
Commander was trained with SAC \cite{sac}, which is a promising RL algorithm for tasks with continuous action space.
We set Commander's number of actions for each episode to 50, and trained for 3M steps.
Meanwhile, Stroker was trained in MuJoCo \cite{mujoco} simulation environment that mimics the real-world experimental setting.
Same as for Commander, SAC was adopted for training Stroker.
Since one Stroker episode performs a single stroke, 1 step per episode was enough to execute the command.
Stroker was trained for 5M steps.

To match the Commander's imaginary canvas—a discrete pixel space—with Stroker's real-world continuous canvas, scaling and rounding are applied.
For instance, our imaginary canvas size is $42\times 42$ pixels and real-world canvas size is $21\times 21$ centimeters, such that 0.5cm in the real-world canvas is translated to 1 pixel in imaginary canvas.
Any distance between 0.25cm and 0.75cm is rounded to 1 pixel.

\begin{figure}[t]
  \centering
  \includegraphics[scale=0.25]{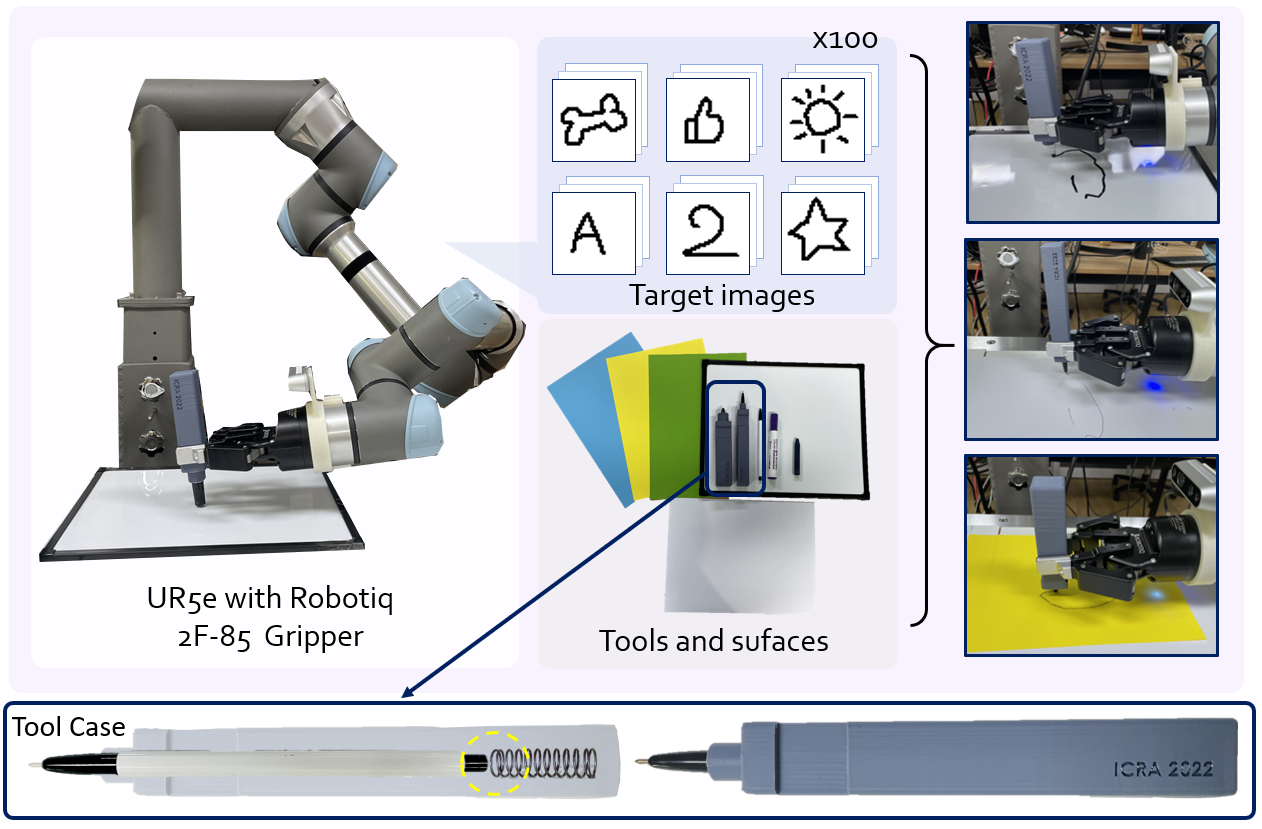}
  \caption{Environmental setup for sketching robot agent. We chose three combinations of drawing tools and surfaces for experiment: whiteboard \& board marker, A4 paper \& pen, and drawing paper \& crayon. A spring-mounted case was used for each tool.}
  \label{setup}
\end{figure}

Fig. \ref{fig:learning} shows that Commander and Stroker successfully learned each task.
Decrease in L2 distance between target images and Commander's drawing indicates that Commander successfully learned to mimic the target image using basic stroke sequences.
Decrease in Euclidean distance and cosine similarity between goal position and reached position of Stroker denotes increase in general performance of Stroker.
On the other hand, Euler angle error of pen rotation decreased slowly because the scale of the related reward term was set lowest.
Fig. \ref{commander_train} shows the Commander's resulting sketches with \textit{Quick, Draw!} as the target images according to the training time.

\subsection{Sketching Agent}
We evaluated our method using UR5e robot with a Robotiq 2F-85 gripper attached as an embodiment for our sketching agent.
The environmental setup for our sketching robot agent is illustrated in Fig. \ref{setup}.
The drawing area is set on the workbench considering the robot's accessibility.
To show the robustness of robot drawing, we considered painting tools and canvases with various thicknesses and roughness.
Board marker, ballpoint pen, and crayon were used as drawing tools.
A4 paper, whiteboard, and colored drawing paper were selected as drawing surfaces.

Protective stop is evoked in many robots including UR5e when the robots are over-stressed from external forces, sometimes even from slight forces caused by the contact of pentip with the surface.
Torque control instead of assigning target joint angles may be adopted to overcome the protection issue, but UR5e does not provide any interfaces about torque control per joint.
Thereby, we used the pen case with a spring to allow our agent to draw freely without significant concern for protective stop (Fig. \ref{setup}).

\begin{figure}[t]
  \centering
  \includegraphics[scale=0.23]{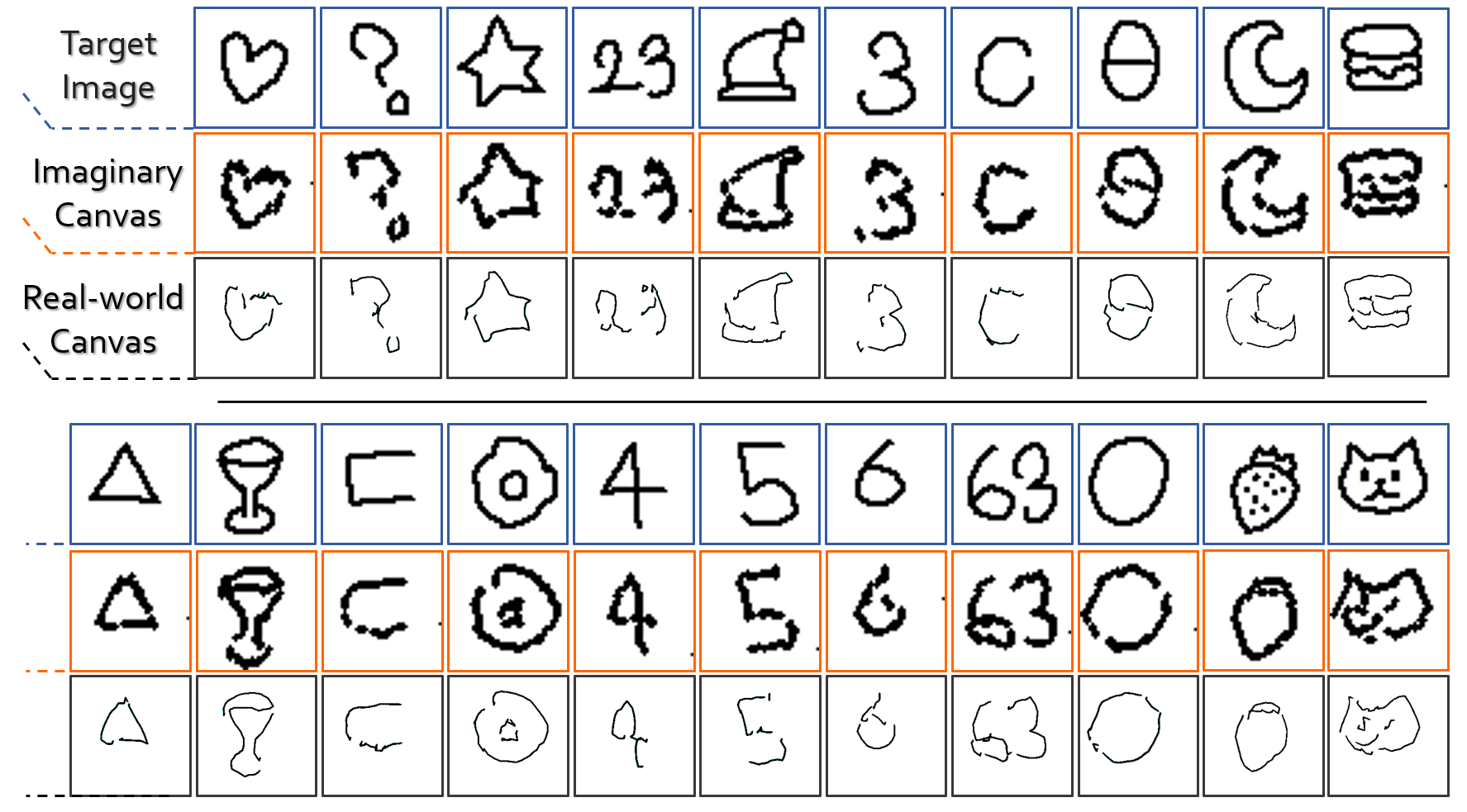}
  \caption{Results on various target images. Commander and Stroker successfully collaborated to perform the target sketches. High similarity between imaginary canvas and real-world canvas shows that synchronization between Commander and Stroker was well-established.}
  \label{result}
\end{figure}

To evaluate our sketching agent, we constructed a hand-drawings dataset including doodles, characters, numbers, and geometries which were not seen in Commander's training.
Fig. \ref{result} shows our agent's attempts at drawing various target images.
Despite encountering the target images for the first time, Commander could successfully decide where to put the strokes sequentially.
There was a tendency to omit details like strawberry seeds, due to the properties of the target images used in training for Commander.
We expect the details can be improved by training Commander with more delicate images.
As Stroker's moves were learned and not completely accurate, the final sketches contained slight errors.

\begin{figure}[t]
  \centering
  \includegraphics[scale=0.23]{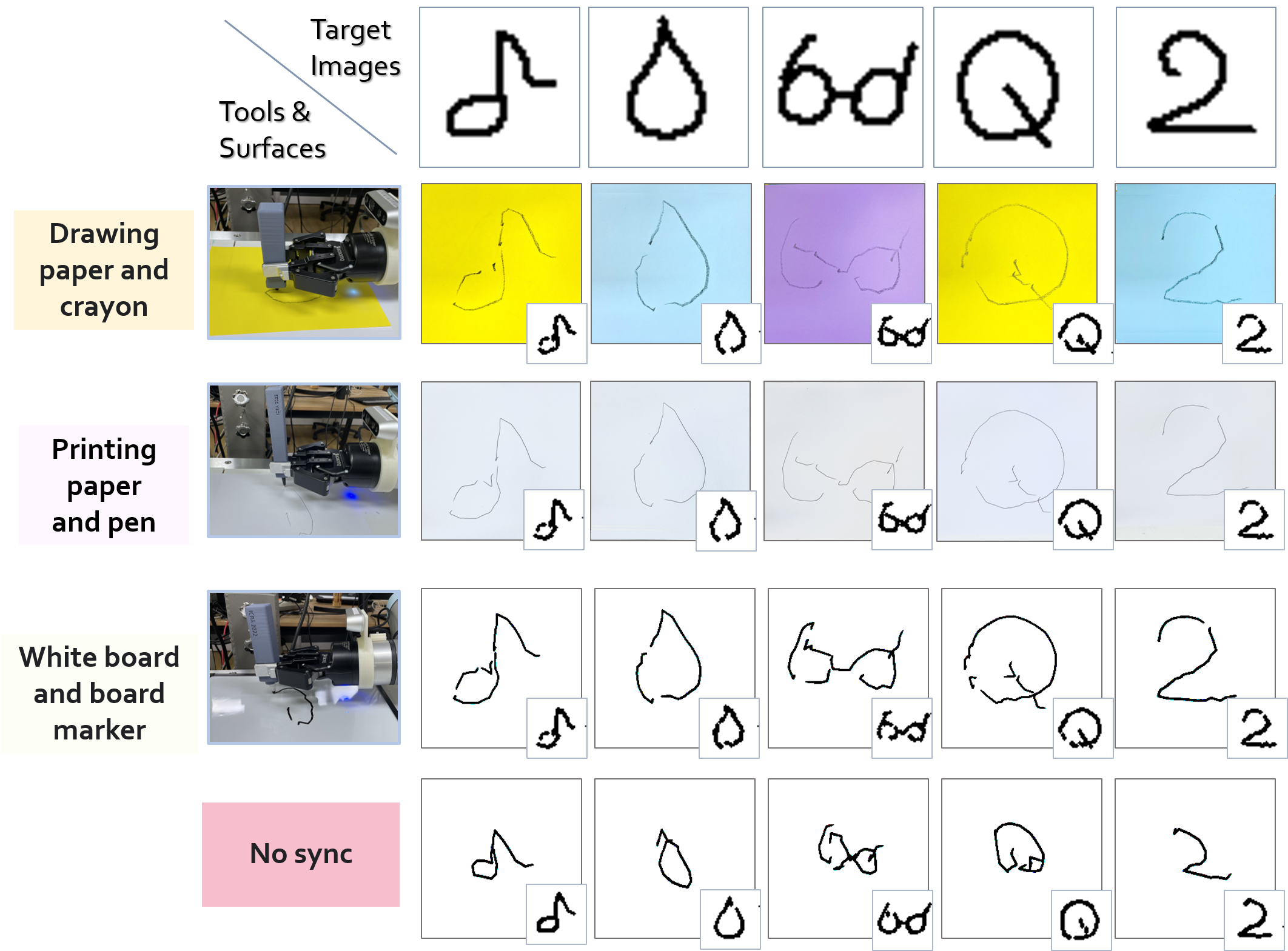}
  \caption{Resulting sketches with varying drawing tools and surfaces. First row is target images. Below three rows are drawing results with three combinations. Small image on the bottom right corner of each real-world sketch is the corresponding imaginary canvas. Final row is the results without synchronizing.}
  \label{result2}
\end{figure}

Fig. \ref{result2} except the final row shows our agent sketching on various surfaces with diverse drawing tools.
Regardless of drawing conditions, the result sketches were almost identical.
We attribute our agent's robustness to our camera-free methodology, which ignores unnecessary visual features such as surface texture, colors, and thicknesses while focusing on the main drawings in imaginary canvas.
Due to the friction differences and kinematic errors, slight differences between each sketch occurred, but these were also well compensated by Commander.
On the other hand, as displayed in the final row of Fig. \ref{result2}, drawing without synchronization technique described in subsection \uppercase\expandafter{\romannumeral3}-E led to impaired sketches, unable to adjust Stroker's errors.

\section{Conclusion and Future Work}
In this paper, we presented a novel deep decoupled hierarchical reinforcement learning framework for training a robot to sketch from scratch without requiring demonstration data, hand-crafted features, inverse kinematics, or camera installation.
Our experimental results on real-world robot drawing tasks showed the effectiveness, stability, robustness, and flexibility of the proposed approach.

We expect our framework to be a good starting point for fully automated training of robotic sketching agents. 
By incorporating with recent AI technologies, the creative and human-friendly sketching agent can be utilized to interact with humans and children for helping and inspiring them.

Although our method does not include real-world visual information for simplification, using a camera as an observation of canvas in the future work could be of great help in the synchronization step to reduce discrepancies when using more diverse drawing tools like brush and spray.




\section*{ACKNOWLEDGMENT}
The authors would like to thank Min Whoo Lee for his careful proofreading.

\bibliographystyle{ieeetr}
\bibliography{ref}

\end{document}